\documentclass{llncs}
\usepackage{graphicx}
\begin{document}

\frontmatter          
\pagestyle{headings}  
\addtocmark{The NUbots Qualification material for RoboCup 2015} 
\mainmatter              
\title{The NUbots Team Description Paper 2015}
\titlerunning{The NUbots Team Description for 2015}  

\author{Josiah Walker
		\and Trent Houliston
		\and Brendan Annable
		\and Alex Biddulph
		\and Jake Fountain
		\and Mitchell Metcalfe
		\and Anita Sugo
		\and Monica Olejniczak
		\and Stephan K. Chalup
		\and Robert A.R. King
		\and Alexandre Mendes
		\and Peter Turner}
\authorrunning{Walker et al.}   
%
\tocauthor{J. Walker,
T. Houliston,
B. Annable,
J. Fountain,
M. Metcalfe,
A. Sugo,
M. Olejniczak,
S. Chalup,
R.A.R. King,
A. Mendes,
P. Turner}

\institute{Newcastle Robotics Laboratory\\ School of Electrical Engineering \& Computer Science\\
Faculty of Engineering and Built Environment\\
The University of Newcastle, Callaghan 2308, Australia\\
Contact: \email{stephan.chalup@newcastle.edu.au}\\
Homepage: \texttt{http://robots.newcastle.edu.au}}

\maketitle              

\begin{abstract}
The NUbots are an interdisciplinary RoboCup team from The University of Newcastle, Australia. The team has a history of strong contributions in the areas of machine learning and computer vision. The NUbots have participated in RoboCup leagues since 2002, placing first several times in the past. In 2014 the NUbots also partnered with the University of Newcastle Mechatronics Laboratory to participate in the RobotX Marine Robotics Challenge, which resulted in several new ideas and improvements to the NUbots vision system for RoboCup. This paper summarizes the history of the NUbots team, describes the roles and research of the team members, gives an overview of the NUbots' robots, their software system, and several associated research projects.

\end{abstract}

\section{Introduction}
The NUbots team, from the University of Newcastle, Australia, competed in the Four Legged League from 2002-2007, within the Standard Platform League from 2008-2011 and subsequently within the Kid-Size Humanoid league since 2012. The NUbots have had a strong record of successes, twice achieving a first place; in 2006 in Bremen, Germany, and, again in 2008 as part of the NUManoid team in Suzhou, China.

The central goal of the NUbots is to compete in RoboCup at a high level by applying current research. The research projects associated with the NUbots team provide unique research and learning opportunities for both undergraduate and postgraduate students in areas related to autonomy and human interaction. Our mission is to contribute to a responsible development and application of robotics. The NUbots also align with the Newcastle Robotics Laboratory which aims to develop and program robots that can support humans not only for routine, challenging, or dangerous tasks, but also to improve quality of life through personal assistance, companionship, and coaching. Some of our projects therefore emphasise anthropocentric and biocybernetic aspects in robotics research~\cite{ChalupOstwald2009,walker2015adaptivemusic,HongEtAl2014,WongEtAl2013}.


\section{Commitment to RoboCup 2015}
The NUbots commit to participation at RoboCup 2015 upon successful qualification. We also commit to provision of a person, with sufficient knowledge of the rules, available as referee during the competition.

\section{History of the NUbots' participation at RoboCup}

The NUbots participated for the first time at RoboCup 2002 in Fukuoka in the Sony Four-Legged League (3rd place). Since then the team has  a strong history of competition and success in the RoboCup SPL/Four-Legged League, obtaining many top three placements and winning the title in 2006 and 2008. In 2013 the NUbots missed out on a place in the quarter finals of the Kidsize League by only one goal. We hope to improve on this in 2015.

The NUbots joined the Kidsize League in 2012 with the DARwIn-OP robots, and ported their SPL codebase to the new platform. The NUbots retained a robust and fast vision and localisation system from the SPL, and ported the B-human NAO walk to the DARwIn-OP for 2012-2013. In 2013 the NUbots presented a curiosity based reinforcement learning approach to gaze planning, which was used with on-line learning for all games during the competition.


\section{Background of the NUbots Team Members}
\begin{itemize}

\item \emph{Brendan Annable} is a 4th year undergraduate honours student studying Software Engineering and is the team's leader. His interests include developing the team's real-time visual debugging systems and other research areas in machine intelligence, computer graphics and GPGPU computing.

\item \emph{Trent Houliston} is studying for a Doctorate of Philosophy in Software Engineering in Software Architecture for Robotics and Artificial Intelligence and is the team's head of research. He designed and implemented the new architecture for the robots, and aided in the development of many of the components.

\item \emph{Jake Fountain} is studying for a Doctorate of Philosophy in Computer Science. Jake has undergraduate degrees in mathematics and science, majoring in physics, with Honours in Computer Science~\cite{fountain2015pointofregard}. His main interests lie in virtual reality and robotics.

\item \emph{Josiah Walker} is studying for a Doctorate of Philosophy in Computer Science in Large Scale Search and Robotics. He works on robot behaviour and machine learning for various NUbot systems. He has been the NUbots team leader for 2013-2014.

\item \emph{Mitchell Metcalfe} is an honours year student studying Computer Science. Mitchell completed undergraduate degrees in Mathematics and Computer Science in 2014. He contributes to the NUbots' localisation system, and is interested in SLAM methods.

\item \emph{Anita Sugo} is a third year undergraduate student studying a combined degree in Mathematics and Science. She is interested in the mathematics used in robotics.

\item \emph{Monica Olejniczak} is a 4th year undergraduate honours student studying Software Engineering. She has contributed to the NUbots' configuration system and is interested in developing debugging tools.

\item \emph{Alex Biddulph} is a 5th year undergraduate student studying Computer Engineering and Computer Science. He is currently working to improve the vision system and develop an alternative controller platform for the Darwin.
He is interested in embedded systems and the melding of software and hardware (electronics) and programmable hardware.

\item \emph{Peter Turner} is technical staff in the School of Electrical Engineering and Computer Science. Peter provides hardware support and assists the team with physical robot design upgrades. 

\item \emph{Dr. Robert King} is a Lecturer in Statistics at the University of Newcastle. His research focus is on flexibly-shaped distributions, statistical computing and Bayesian knowledge updating. He joined the NUbots in 2004 and has developed a special interest in the RoboCup rules and refereeing.

\item \emph{Dr. Alexandre Mendes} is deputy head of the Newcastle Robotics Lab. He is a Senior Lecturer in Computer Science and Software Engineering. He joined the group in September 2011 and his research areas are algorithms and optimisation.

\item \emph{A/Prof. Stephan Chalup} is the head of the Newcastle Robotics Lab. He is an Associate Professor in Computer Science and Software Engineering. He is one of the initiators of the University of Newcastle's RoboCup activities since 2001. His research area is machine learning and anthropocentric robotics.
\end{itemize}
We also acknowledge the valuable input of colleagues from the Newcastle Robotics Laboratory, team members of previous years
and the Interdisciplinary Machine Learning Research Group (IMLRG) in
Newcastle, Australia. Details are linked to the relevant webpages at
\texttt{www.robots.newcastle.edu.au}.

\section{Hardware and Software Overview}
The NUbots use the DARwIn-OP robot with foot sensors. The team has seven of these robots that are of the standard design with the exception of a slightly reduced foot size. The team also hopes to field modified DARwIn-OP robots consisting of a full HD camera, an ODroid-XU computer and an updated motor communications board as a part of a student project. 


The NUbots team's major research focus is on using machine learning methods within the software systems of the robot to achieve increased performance and autonomy~\cite{ChalupEtAlSMC2007}. The current NUbots software source is available from \cite{nubotsGit} and is covered under the GPL. This code includes associated toolkits for building and deploying the software. Our software is designed to work on multiple robotic platforms, and all of the individual modules have been designed to be easily used in other systems. The flexibility of our approach has been demonstrated in a deployment of the NUbots vision system on a marine platform \cite{renton2014robotx}. 

Following development of a new software system in 2014, the NUbots are now focusing on current and emerging challenges within the RoboCup Kid-size League. These include robust, adaptable image segmentation; generic ball detection; and improving the architecture of current walk engines to allow for more rapid experimentation and improvement. The NUbots software is designed to allow new teams and team members to easily understand and innovate on existing code, and is made freely available to encourage research and innovation.




\section{Acknowledgement of Use of Code}
The NUbots DARwIn-OP robots use a walk engine based on the 2013 Team Darwin code release. We acknowledge the source of this code. The NUbots have ported this code to C++ and restructured the logic, making numerous structural and technical changes since. 

\section{Hardware Enhancements since RoboCup 2014}
At RoboCup 2014 we trialled rapid prototyping for a new head design for the Darwin-OP robots to fit upgraded Logitech C920 cameras. Since this time we have been improving designs and readying for an open source release once remaining issues are resolved. We field these otherwise standard Darwin-OP robots under the name NU-Darwin.

We have been partnering with Kontron Australia to develop more powerful embedded pc boards in order to upgrade our capabilities and deploy new robotics platforms. This upgrade will see higher quality accelerometers and gyroscopes and more hardware communications channels added to the robots, as well as an upgrade to a quad-core celeron platform with access to OpenCL.






\section{Research Areas}

\noindent\textbf{Robot Vision:} Vision is one of the major research areas associated with the Newcastle Robotics Laboratory. Several subtopics have been investigated including object recognition, horizon determination, edge detection, model fitting and colour classification using ellipse fitting, convex optimisation and kernel machines. Recent work has resulted in a fully-autonomous method of colour look-up table adaptation for changing lighting conditions, allowing us to overcome one of the major limitations of the colour look-up table system. Publications are available e.g. from~\cite{budden2012colour,budden2012ball,henderson_2007,nickin_2007,NUBOT2006,Henderson2008,flannery2013ransac,budden2013salient,renton2014robotx}.

\noindent\textbf{Localisation and Kalman Filters:} Research on the
topic of localisation focused on Bayesian approaches to robot
localisation including Kalman Filter and particle filter
based methods. We are interested in
modifications of the Kalman Filter to handle non-ideal information
from vision, incorporate increased information from multiple agents,
and effectively utilise non-unique objects.

\noindent\textbf{Development of the Robot Bear:} In a collaborative effort with the company Tribotix and colleagues in design, a bear-like robot (called Hykim) was developed~\cite{ChalupEtAl2006}. It has a modular open platform using Dynamixel servos.

\noindent\textbf{Biped Robot Locomotion:} The improvement of walking speed and stability has been investigated by the NUbots for several years and on different platforms: On the AIBO robot we achieved one of the fastest walks at that time by walk parameter evolution \cite{QuinlanEtAlACRA2003,ChalupEtAlSMC2007}. On the Nao robot we improved existing walk engines by modifying the joint stiffnesses, or controller gains, \cite{Kulk2008,Kulk2010,Kulk2010a}. The stiffnesses were selected through an iterative process to maximise the cost of transport. We investigated the application of Support Vector Machines and Neural Networks to proprioception data for sensing perturbations during pseudo quiet stance. Walk improvements have been primarily done via optimisation techniques \cite{Kulk2011a},  
with recent improvements to our framework for online optimisation of bipedal humanoid locomotion.
The use of spiking neural networks has been trialled in simulation~\cite{WiklendtChalup2008}. Prior to RoboCup 2012 the walk engine developed by the leading SPL team BHuman~\cite{BHumanWalk2010} was ported to the DARwIn-OP platform, and a variety of optimisation techniques were developed and successfully applied to improve walking speed and stability of the DARwIn-OP`\cite{budden2013probabilistic}. Multi-agent walk optimisation is being developed for this year's competition.

\noindent\textbf{Reinforcement Learning, Affective Computing and Robot Emotions:} We investigate the feasibility of reinforcement learning or neurodynamic programming for applications such as motor control and music composition. Concepts for affective computing are developed in multidisciplinary projects in collaboration with the areas of architecture and cognitive science. The concept of emotion is important for selective memory formation and action weighting and continues to gain importance in the robotics community, including within robotic soccer. A number of projects in the Newcastle Robots Laboratory already address this topic~\cite{Pareidolia2010,HongEtAl2013a,HongEtAl2013b,WongEtAl2013,HongEtAl2014}.

\noindent\textbf{Gaze analysis and head movement behavioural learning:} We investigated methods for human and robot pedestrian gaze analysis in~\cite{JalalianEtAl_CAADRIA2011,WongEtAl2012} as well as space perception, way finding and the detection and analysis of salient regions~\cite{BhatiaChalup2013,BhatiaEtAl2013,BhatiaChalupOstwald2013}. Recently we  applied motivated reinforcement learning techniques to optimising head movement behaviour, providing a robust algorithm by which a robot learns to choose landmarks to localise efficiently during a soccer game~\cite{FountainEtAl2014}. This algorithm was used in competition at RoboCup 2013, and is in the process of being adapted for our new software architecture.

\noindent\textbf{Manifold Learning:} In several projects we
investigate the application of non-linear dimensionality reduction
methods in order to achieve more understanding of, and more precise
and efficient processing of, high-dimensional visual and acoustic data.
\cite{ChalupEtAl2007b,WongEtAl2012}.

\noindent\textbf{Other new projects:} Much work has been focused on the underlying software architecture and external utilities to enable flexibility and extensibility for future research. Projects undertaken include improving the configurability of the software system via real-time configuration updates, development of a web-based online visualisation and debugging utility \cite{AnnableEtAl2014} and the application of software architectural principles to create a multithreaded event-based system with almost no run-time overhead. Some of this work is still in progress by new undergraduate students who have joined the team.

\section{Related Research Concentrations}
The \emph{Interdisciplinary Machine Learning Research Group (IMLRG)} investigates different aspects of machine learning and data mining in theory, experiments and applications. The IMLRG's research areas include: Dimensionality reduction, vision processing, robotics control and learning,  evolutionary computation, optimisation, reinforcement learning, and kernel methods. Links to publications can be found at the NUbots' webpage
\texttt{http://robots.newcastle.edu.au/}

\bibliographystyle{plain}
\bibliography{nubots}

\end{document}